\documentclass{article}
\PassOptionsToPackage{numbers,sort&compress}{natbib}
 \usepackage[preprint]{neurips_2026}

\usepackage[utf8]{inputenc} 
\usepackage[T1]{fontenc}    
\usepackage{hyperref}       
\usepackage{url}            
\usepackage{booktabs}       
\usepackage{amsfonts}       
\usepackage{nicefrac}       
\usepackage{microtype}      
\usepackage{xcolor}         
\usepackage{amsmath}
\usepackage{graphicx}
\usepackage{multirow}
\usepackage{algorithm}
\usepackage{algorithmic}
\usepackage{amsthm}
\usepackage{amssymb}
\usepackage{tabularx}
\usepackage{threeparttable}
\usepackage{ragged2e}

\title{Normative Loss Landscape Navigation: A Trajectory-Based Approach to Mitigating Forgetting in Incremental Learning}

\author{%
  Isabelle Aguilar\thanks{Corresponding author} \\
  School of Biomedical Engineering\\
  University of Sydney\\
  Sydney, NSW 2050 \\
  \texttt{isabelle.aguilar@sydney.edu.au} \\
  \And
  Zayn Andre Zainal \\
  School of Biomedical Engineering \\
  University of Sydney\\
  Sydney, NSW 2050 \\
  \texttt{andre.zainal@sydney.edu.au} \\
    \And
  Luis Fernando Herbozo Contreras \\
  School of Biomedical Engineering \\
  University of Sydney\\
  Sydney, NSW 2050 \\
  \texttt{luis.herbozocontreras@sydney.edu.au} \\
      \And
  Zhaojing Huang \\
  School of Biomedical Engineering \\
  University of Sydney\\
  Sydney, NSW 2050 \\
  \texttt{zhaojing.huang@sydney.edu.au} \\
  \And
  Omid Kavehei \\
  School of Biomedical Engineering \\
  University of Sydney\\
  Sydney, NSW 2050 \\
  \texttt{omid.kavehei@sydney.edu.au} \\
}

\begin{document}

\maketitle

\begin{abstract}
Continual learning models suffer from \textit{catastrophic forgetting} when trained sequentially on non-stationary data distributions. Previously, this has been addressed through weight regularization. While preconditioning gradients offer a promising alternative to mitigate forgetting, current approaches are myopic. Conversely, standard regularization methods apply rigid, scalar Euclidean penalties that entirely ignore the underlying Riemannian geometry of the parameter space. To overcome this gap, we propose TMLN (\textbf{T}rajectory-\textbf{M}odulatory \textbf{L}andscape \textbf{N}avigation), a normative navigation policy that formalizes continual learning as an optimal control problem over a curved loss landscape. TMLN utilizes a memory-efficient diagonal empirical Fisher Information Matrix (FIM) to define a localized Riemannian manifold. To compensate for the spatial limitations of the diagonal approximation, TMLN dynamically modulates a preconditioner using the normalized historical trajectory of the network's parameter values. By integrating this trajectory-based preconditioning directly into the gradient update, we actively shield historically critical parameter directions without relying on additive penalties. Empirical evaluations on class- and domain-incremental benchmarks demonstrate that our method significantly reduces the loss barrier between consecutive tasks.
\end{abstract}

\section{Introduction}
\label{sec:intro}
Deep neural networks achieve remarkable performance on stationary datasets but fail when tasked with learning sequentially from non-stationary data streams. This failure manifests itself as the stability-plasticity dilemma, where networks must remain plastic enough to acquire new knowledge while stable enough to retain previously learned representations. In standard gradient-based optimization, sequential learning can lead to one of two extremes. The most prominent is \textit{catastrophic forgetting} \cite{mccloskey1989catastrophic, french1999catastrophic}, where shortsighted gradient updates for a new task overwrite the crucial parameter configurations for previous tasks. In contrast, attempts to rigidly protect these parameters can induce \textit{intransigence} \cite{li2023fixed, chaudhry2018riemannian}, where the model's capacity to adapt to new distinct tasks is incapacitated. Solving continual learning requires navigating the delicate balance between these potential outcomes.

Current methods address the stability-plasticity dilemma by using either structural memory buffers or regularization. However, these approaches rely on flawed geometric assumptions. Foundational regularization techniques (largely inspired by biological synaptic consolidation) attempt to anchor weights to past distributions. However, applying diagonal approximations as rigid, static penalties at discrete points in time eventually freezes the local geometry, artificially inducing intransigence and losing plasticity on new tasks over time \cite{van2025computation, dohare2024loss}. State-of-the-art replay and optimization methods recover some of this geometry but introduce extreme memory overheads or heuristic Euclidean measures that lack formal optimization guarantees. Furthermore, bio-inspired learning rules show promise, but struggle to scale to high-dimensional deep architectures \cite{bartunov2018assessing}.

\begin{figure}[htpb!]
\centering
\includegraphics[width=0.99\textwidth]{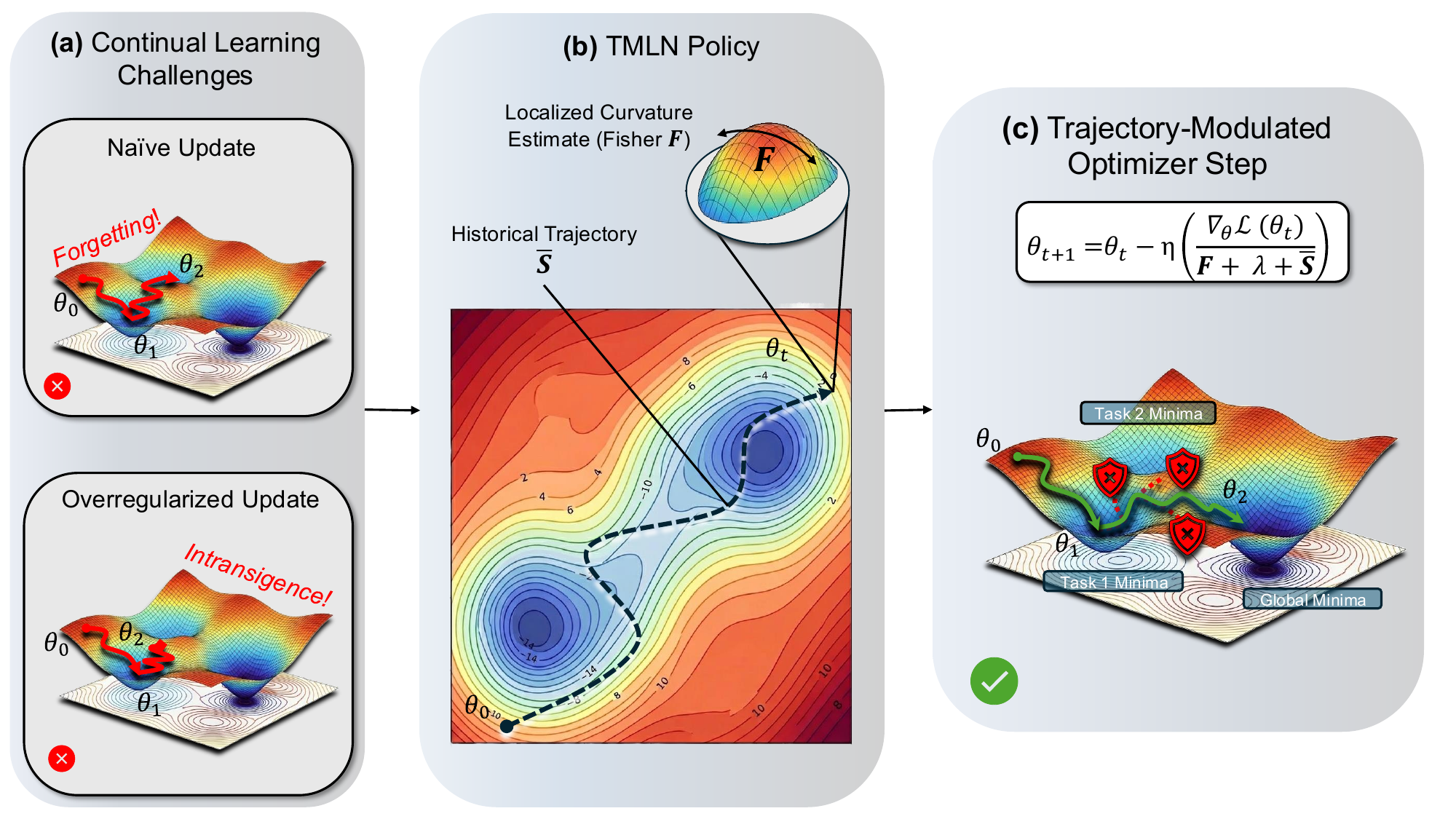}
\caption{\textbf{TMLN Framework.} \textbf{(a)} On a loss landscape, continual learning tackles both forgetting (in which the model moves out of task 1's local minima and into task 2's local minima) and intransigence (where the model struggles to move out of task 1's minima). \textbf{(b)} \textit{Trajectory Modulated Landscape Navigation} measures local curvature and historical trajectory, and incorporates it into a preconditioned gradient step. \textbf{(c)} Our method efficiently modulates optimization learning rules, allowing sequential learning while preventing movement in steep directions that harm past knowledge. This figure is produced with the aid of Gemini 3.}
\label{fig:fig1} 
\end{figure}

We argue that continual learning must be fundamentally recast from discrete, constrained gradient steps into a continuous geometric problem \cite{vastola2025gradient}. In this study, we elevate the biological concept of synaptic consolidation into a formal mathematical framework by \textbf{framing sequential learning as an optimal control problem over a Riemannian manifold}. Our approach, denoted as \textit{Trajectory Modulated Landscape Navigation} (TMLN), optimizes a learning trajectory that intrinsically respects the curvature of the network's parameter space. As shown in Fig.~\ref{fig:fig1}, by navigating the geometry of the loss landscape, we mathematically bound the update rules to prevent forgetting while maximizing plasticity in unconstrained directions.

In this work, our \textbf{primary contributions} are as follows:
\begin{itemize}
    \item We formulate a novel optimization objective that frames sequential continual learning as trajectory optimization over a Riemannian manifold.
    \item We derive a memory-efficient metric tensor approximation that captures the local loss landscape geometry.
    \item We empirically demonstrate superior stability-plasticity trade-offs across rigorous single-head benchmarks, Split CIFAR-100, and CORe50.
\end{itemize}

\section{Related Work}
\label{sec:related_work}
We contextualize our framework within the intersecting domains of regularization in continual learning, biologically-inspired continual learning methods, and Riemannian optimization.

\subsection{Regularization \& Optimization Strategies}
\label{subsec:reg_opt}
There are typically three main strategies for continual learning: (1) regularization \cite{kirkpatrick2017overcoming, li2017learning}, (2) replay \cite{rolnick2019experience, shin2017continual, wang2024comprehensive}, and (3) architectural-based/modulation \cite{de2021continual, mallya2018packnet, yan2021dynamically}. Regularization-based continual learning mitigates catastrophic forgetting by anchoring network weights to a prior distribution based on previous tasks. Foundational baselines such as Elastic Weight Consolidation (EWC) \cite{kirkpatrick2017overcoming}, Synaptic Intelligence (SI) \cite{zenke2017continual}, and Memory Aware Synapses (MAS) \cite{aljundi2018memory} formulate this by penalizing gradient updates in highly sensitive parameter directions. However, to maintain computational tractability, these methods critically rely on strictly diagonal approximations of the Fisher Information Matrix (FIM) \cite{van2025computation} or on empirical path-integral point-wise approximations. This static application artificially flattens the rich, continuous structure of the loss landscape into a rigid space, failing to capture how curvature evolves safely along a training trajectory.

Recent state-of-the-art optimization strategies attempt to overcome these limitations but introduce secondary bottlenecks. For example, Online Curvature-Aware Replay (OCAR) \cite{urettini2025online} leverages a Kronecker-factored (K-FAC) block-diagonal approximation of the FIM to precondition gradients, bridging second-order optimization with KL-divergence constraints. Alternatively, methods such as Utility-based Perturbed Gradient Descent (UPGD) \cite{elsayed2023utility} attempt to avoid catastrophic forgetting by injecting stochastic noise scaled by parameter utility, effectively bumping the optimization trajectory out of high-forgetting regions. Although these approaches attempt to step beyond naive diagonal approximations, they remain flawed for scalable continual learning. OCAR requires memory-intensive, interleaved replay buffers to stabilize FIM estimation, whereas UPGD relies on heuristic, Euclidean stochasticity rather than formally accounting for the intrinsic curvature of the loss landscape.

\subsection{Biological Approaches}
\label{subsec:bio_approaches}
A parallel lineage of research attempts to mitigate catastrophic forgetting by mimicking the neurobiological mechanisms of the mammalian brain. Strategies in this domain heavily leverage metaplasticity \cite{jedlicka2022contributions, aguilar2025continuous, benna2016computational}, localized Hebbian plasticity \cite{oja1982simplified}, or energy-efficient neuromorphic computing paradigms such as Spiking Neural Networks (SNNs) \cite{davies2018loihi, eshraghian2023training}. These approaches conceptualize memory retention through phenomenological models of complex synapses, where learning rates are dynamically modulated based on the historical importance of localized parameters. However, this domain is characterized by a critical limitation: a severe lack of normative, mathematically rigorous optimization guarantees. Although empirically interesting and hardware-efficient, local learning rules and heuristic synaptic consolidation are largely disconnected from the global geometry of the network's loss landscape \cite{yang2021taxonomizing}. Consequently, these models may struggle to scale to complex, high-dimensional deep architectures, because they rely on local update rules rather than explicitly addressing the global optimization geometry of the loss landscape \cite{erhan2009difficulty, fort2020deep}. 

\subsection{Riemannian Geometry \& Natural Gradients}
\label{subsec:geometry}
Optimization in deep neural networks is fundamentally a geometric problem. Natural Gradient Descent (NGD) \cite{rattray1998natural} established that moving along the steepest descent direction in the Riemannian manifold (with the FIM defining the metric tensor) yields optimal parameter updates. Tractable approximations, such as Kronecker-factored Approximate Curvature (K-FAC) \cite{martens2015optimizing}, have successfully scaled NGD to single-task architectures in modern deep learning. However, translating Riemannian optimization to non-stationary, sequential data streams exposes severe computational and mathematical bottlenecks. 

Standard regularization methods treat the parameter space as a Euclidean domain, and oversimplify the true topology of neural network loss landscapes \cite{fort2019large, ly2025optimization}. Recently, connections between catastrophic forgetting and the geometric properties of the loss landscape have been established; several studies \cite{mirzadeh2020understanding, lyle2023understanding, bian2024make} have provided theoretically robust reasoning, supported by empirical evidence, that converging to wider local minima significantly mitigates forgetting in sequential task learning. However, translating these insights into action naturally points to Riemannian optimization, which exposes severe bottlenecks in non-stationary environments. Naively applying natural gradients requires explicitly maintaining, recursively updating, and inverting the metric tensor throughout the task sequence. With this approach, a costly amount of memory is required to store historical task curvatures; otherwise, catastrophic forgetting of the manifold's geometry occurs. In the latter case, the network loses the structure of the safe parameter space even before the functional mappings themselves are degraded \cite{kao2021natural}. 

\paragraph{Bridging the Conceptual Gap.} Scalable continual learning requires both the geometric precision of natural gradients and the computational efficiency of localized heuristics. We address the limitations of existing techniques by navigating the geometry of the loss landscape without relying on explicit metric tensor inversions or stochastic Euclidean noise. Our approach safely navigates previously preserved Riemannian curves. This framework successfully scales the theoretical rigor of information geometry to high-dimensional architectures, definitively addressing the stability-plasticity dilemma.

\section{Trajectory-Modulated Landscape Navigation (TMLN)}
\label{sec:methods}

\paragraph{Continual Learning as Landscape Navigation.}
We frame continual learning not merely as a regularization problem, but as an optimal control problem over a partially observable loss landscape, similar to \cite{vastola2025gradient}. Standard gradient descent rules navigate the loss landscape blindly, often destroying the flat minima of previous tasks (inducing catastrophic forgetting) or becoming trapped by rigid penalties (causing intransigence). 

To address this, we introduce Trajectory Modulated Landscape Navigation (TMLN). TMLN actively bends the optimization trajectory of the current task by preconditioning the gradient. However, rather than relying on computationally prohibitive full-matrix second-order approximations, we compute a memory-efficient diagonal curvature estimate strictly from the sampling buffer. To compensate for the spatial correlations lost by the diagonal approximation, we introduce a novel dynamic damping mechanism modulated by the historical trajectory of the parameters.

\subsection{Curvature Estimation via Buffer Replay}
\label{subsec:FIM}
Let $\theta \in \mathbb{R}^D$ be the parameters of the neural network. To navigate the loss landscape, we require an estimate of the local geometry of previous tasks via Eq.~\ref{eq:fisher_diag}. We use a small episodic memory buffer $\mathcal{M}$ as our spatial sensor. 

To prevent the curvature estimate from being biased by the current task's trajectory, the geometry is evaluated strictly at the task boundary $T_{k-1} \to T_k$. We compute the empirical FIM metric $\mathbf{F} \in \mathbb{R}^D$ using only the samples preserved in $\mathcal{M}$:

\begin{equation}
F_i = \frac{1}{|\mathcal{M}|} \sum_{(x, y) \in \mathcal{M}} \left( \nabla_{\theta_i} \log p(y | x; \theta) \right)^2
\label{eq:fisher_diag}
\end{equation}

\subsection{Trajectory-Modulated Damping}
\label{subsec:traj}
By invoking the mean-field assumption (using a diagonal $\mathbf{F}$), we sacrifice the off-diagonal covariance between parameters. In deep networks, this can falsely indicate that certain highly-correlated parameter directions are safe to traverse \cite{van2025computation}. 

To compensate for this lack of spatial information and extend the optimization horizon, we introduce a temporal constraint, which we term trajectory-modulated damping. We track the continuous path integral \cite{chaudhry2018riemannian} of the parameters to measure their historical contribution to loss reduction. During task $k$, the parameter importance $\omega_i^{(k)}$ is accumulated dynamically:

\begin{equation}
\omega_i^{(k)} = - \int_{t_0}^{t_1} \nabla_{\theta_i} \mathcal{L}^{(k)}(\theta(t)) \cdot \frac{d\theta_i}{dt} dt \approx - \sum_{t} \nabla_{\theta_i} \mathcal{L}_t \cdot \Delta \theta_{i,t}
\label{eq:path_integral}
\end{equation}

When an update direction consistently reduces loss, $(-\nabla_{\theta_i} \mathcal{L}_t\Delta\theta_{i,t})$ is positive; we clamp negative values (via $\mathrm{ReLU}$) to avoid rewarding parameters that increase loss or oscillate (see Algorithm \ref{alg:tmln}, line 17).

At the end of task $k$, we normalize this trajectory (computed in Eq.~\ref{eq:path_integral}) against the Riemannian distance traveled, yielding the parameter-specific sensitivity score $S_i^{(k)}$ (Eq.~\ref{eq:sensitivity_score}):

\begin{equation}
S_i^{(k)} = \frac{\omega_i^{(k)}}{\frac{1}{2} F_i (\Delta \theta_i^{(k)})^2}
\label{eq:sensitivity_score}
\end{equation}

To construct our dynamic damping tensor, we maintain a normalized running history of these trajectory scores for all tasks seen: $\bar{S}_i = \sum_{j=1}^k \left( S_i^{(j)} / \max_m S_m^{(j)} \right)$. This ensures that parameters historically vital for optimization are aggressively shielded, even if the diagonal Fisher incorrectly predicts a low local curvature.

\subsection{The TMLN Optimization Step}
\label{subsec:tmln_opt}
We unify the spatial curvature ($\mathbf{F}$) and the temporal trajectory ($\bar{S}$) into a single preconditioned optimization step in Eq.~\ref{eq:tmln_update}. For a given data batch at training step $t$, the network is updated via:

\begin{equation}
\theta_{t+1} = \theta_t - \eta \left( \frac{\nabla_{\theta} \mathcal{L} (\theta_t)}{\mathbf{F} + \lambda\mathbf{1} + \gamma\bar{S}} \right)
\label{eq:tmln_update}
\end{equation}

where $\eta$ is the learning rate, $\lambda$ is the base Tikhonov damping to ensure numerical stability, and $\gamma$ modulates the temporal trajectory strength; the significance of this regularization is made clear in Appendix \ref{sec:TMMT_construction}. Note that the division in Eq.~\ref{eq:tmln_update} denotes element-wise division, and $\mathbf{1} \in \mathbb{R}^D$ is a vector of ones. By pushing the regularization directly into the denominator of the gradient update, TMLN bypasses the Euclidean gradient conflicts inherent in additive penalty methods. The complete TMLN training protocol is outlined in pseudocode Algorithm \ref{alg:tmln}.

\begin{algorithm}[htp!]
\caption{Trajectory-Modulated Landscape Navigation (TMLN)}
\label{alg:tmln}
\begin{algorithmic}[1]
\REQUIRE Sequence of task datasets $\mathcal{T} = \{\mathcal{D}_1, \mathcal{D}_2, \dots, \mathcal{D}_K\}$, Epochs $E$, Learning rate $\eta$, Base damping $\lambda_{\text{base}}$, Buffer capacity $M$
\STATE \textbf{Initialize}: Network parameters $\boldsymbol{\theta}$, Buffer $\mathcal{M} \leftarrow \emptyset$
\STATE \textbf{Initialize State Variables}: Diagonal Fisher $\mathbf{F} \leftarrow \mathbf{0}$, Sensitivity Score History $\bar{\mathbf{S}} \leftarrow \mathbf{0}$, Path Integral $\boldsymbol{\omega} \leftarrow \mathbf{0}$, Anchor $\boldsymbol{\theta}_{\text{prev}} \leftarrow \boldsymbol{\theta}$

\FOR{each task $k = 1, \dots, K$}
    \FOR{each epoch $e = 1, \dots, E$}
        \FOR{each mini-batch $\mathcal{B}_k \sim \mathcal{D}_k$}
            \STATE Sample replay batch $\mathcal{B}_{\mathcal{M}} \sim \mathcal{M}$ 
            \STATE $\tilde{\mathcal{B}} \leftarrow \mathcal{B}_k \cup \mathcal{B}_{\mathcal{M}}$ 
            
            \STATE $\mathcal{L} \leftarrow \frac{1}{|\tilde{\mathcal{B}}|} \sum_{(\mathbf{x}, y) \in \tilde{\mathcal{B}}} \ell(f_{\boldsymbol{\theta}}(\mathbf{x}), y)$ 
            \STATE $\mathbf{g} \leftarrow \nabla_{\boldsymbol{\theta}} \mathcal{L}$ 
            
            \STATE $\tilde{\mathbf{g}} \leftarrow \mathbf{g} \oslash \left(\mathbf{F} + \lambda_{\text{base}}\mathbf{1} + \bar{\mathbf{S}}\right)$ \COMMENT{Precondition with modulated damping}
            
            \STATE $\Delta \boldsymbol{\theta} \leftarrow -\eta \tilde{\mathbf{g}}$ 
            \STATE $\boldsymbol{\theta} \leftarrow \boldsymbol{\theta} + \Delta \boldsymbol{\theta}$ \COMMENT{Parameter update}
            \STATE $\boldsymbol{\omega} \leftarrow \boldsymbol{\omega} - (\mathbf{g} \odot \Delta \boldsymbol{\theta})$ \COMMENT{Accumulate path integral}
        \ENDFOR
    \ENDFOR
    
    \STATE Update buffer $\mathcal{M}$ with samples from $\mathcal{D}_k$ 
    
    \STATE $\mathbf{s}^{(k)} \leftarrow \text{ReLU}(\boldsymbol{\omega}) \oslash \left( \frac{1}{2} \mathbf{F} \odot (\boldsymbol{\theta} - \boldsymbol{\theta}_{\text{prev}})^{\odot 2} + \epsilon \mathbf{1} \right)$
    \STATE $\bar{\mathbf{S}} \leftarrow \bar{\mathbf{S}} + \frac{\mathbf{s}^{(k)}}{\max(\mathbf{s}^{(k)})}$ 
    
    \STATE $\mathbf{F} \leftarrow \frac{1}{|\mathcal{M}|} \sum_{(\mathbf{x},y) \in \mathcal{M}} \left( \nabla_{\boldsymbol{\theta}} \log p(y|\mathbf{x}; \boldsymbol{\theta}) \right)^{\odot 2}$ 
    
    \STATE $\boldsymbol{\theta}_{\text{prev}} \leftarrow \boldsymbol{\theta}$ 
    \STATE $\boldsymbol{\omega} \leftarrow \mathbf{0}$
\ENDFOR
\end{algorithmic}
\end{algorithm}

\subsection{Computational and Memory Complexity}
\label{subsec:complexity}
A fundamental limitation of applying true Riemannian optimization (e.g., Natural Gradient Descent) to deep continual learning is the excessive cost of explicitly computing and inverting the FIM \cite{soen2024trade}. For a neural network parameterized by $\theta \in \mathbb{R}^N$, storing the full dense FIM requires $\mathcal{O}(N^2)$ memory, and computing its inverse scales with $\mathcal{O}(N^3)$ time complexity. This is intractable for modern deep architectures.
In contrast, TMLN achieves the geometric benefits of Riemannian navigation while strictly maintaining linear complexity. More in-depth analysis can be found in Appendix ~\ref{appendix:overhead}.

\paragraph{Memory Overhead.} TMLN relies on a diagonal approximation of the curvature, supplemented by the historical trajectory state variables. We explicitly store the diagonal FIM, the parameter anchor points of the previous task, and the running trajectory buffer (the normalized path integral $\pi$). Because these are all parameter-wise vectors, the total memory overhead of the optimizer is strictly $\mathcal{O}(N)$. If combined with a small episodic buffer of parameter-equivalent memory of $B$, the total memory footprint becomes $\mathcal{O}(N + B)$, which remains vastly superior to the scaling $\mathcal{O}(N^2)$ of explicit second-order methods.
\paragraph{Computational Cost.} By preconditioning the gradient updates using a modulated diagonal formulation rather than a full tensor, the intractable $\mathcal{O}(N^3)$ matrix inversion is reduced to a simple $\mathcal{O}(N)$ element-wise division. The trajectory modulation updates (which include calculating the Euclidean displacement and accumulating the path integral) are also strictly element-wise operations performed once per step or task boundary. Consequently, the wall-clock training time of TMLN is asymptotically equivalent to that of standard first-order methods such as SGD and localized heuristics such as EWC. 

\section{Experiments}
\label{sec:experiments}

\subsection{Experimental Setup}
\label{subsec:exp_setup}
All experiments follow a single-head continual learning setting, meaning that in a stream of \textit{k} tasks, a task identifier or label was unknown during evaluation \cite{chaudhry2018riemannian}. This is a more challenging and realistic setting than multi-head settings, where tasks have separate classifier heads or a task label is given during test time \cite{zhou2026class, kumar2024meta}. In our setup, we also avoid linear growth of task-specific modules, allowing scalability of our strategy.

\paragraph{Benchmarks.} 
We adopt the class-incremental Split CIFAR-100 \cite{krizhevsky2009learning, zenke2017continual} and domain-incremental CORe50 \cite{lomonaco2017core50} datasets for evaluation. Split CIFAR-100 consists of $10$ tasks, each with $10$ classes. CORe50 consists of $8$ total tasks, each with $10$ classes. Additionally, we use Split CIFAR-10 \cite{krizhevsky2009learning, zenke2017continual} for an analytical study on the stability-plasticity tradeoff. Each method is evaluated over five runs, and we report the average accuracy of all tasks at the end of training. Average forgetting and intransigence are also reported as representative measures of the stability and plasticity, respectively \cite{chaudhry2018riemannian}.

\paragraph{Baselines.} 
We compare TMLN with foundational baselines and state-of-the-art continual learning strategies. We use a standard SGD optimizer with momentum as a Naive baseline. We consider the baseline ER \cite{rolnick2019experience} as a standard experience replay method. Additionally, we compare our method with the baseline regularization strategy EWC++ \cite{chaudhry2018riemannian, kirkpatrick2017overcoming} and the state-of-the-art RWalk \cite{chaudhry2018riemannian}. We also compare with a recent state-of-the-art method, OCAR \cite{urettini2025online}, which combines optimization and replay techniques. Finally, we evaluate on a state-of-the-art continual learning optimizer that targets both plasticity and stability, UPGD \cite{elsayed2023utility}.

\paragraph{Metrics.}
To evaluate continual learning ability, we report Average Accuracy (\textit{ACC}), which measures the mean performance across all sequential tasks after the whole training sequence is completed, as defined in \cite{lopez2017gradient}. Additionally, we report Average Forgetting (\textit{FM}), and Intransigence (\textit{INT}) metrics formalized in \cite{chaudhry2018riemannian}. These measures quantify the underlying stability-plasticity balance that defines continual learning. Forgetting quantifies the degradation in performance on previous tasks from their peak accuracy, directly measuring the network's stability and susceptibility to interference. Intransigence measures the model's inability to acquire new knowledge compared to a standard oracle, serving as a strict reference for the loss of plasticity. Together, these metrics provide a comprehensive evaluation of how effectively a method navigates the optimization landscape \cite{diaz2018don}.

\paragraph{Architecture.} 
We use a simplified ResNet18 \cite{he2016deep} as the backbone for all CIFAR experiments and a MobileNetV2 \cite{sandler2018mobilenetv2} for CORe50 experiments. All models are trained from scratch, with no pretraining involved. 

\paragraph{Training.}
With the exception of UPGD, all evaluated methods strictly employ standard Stochastic Gradient Descent (SGD). Curvature-aware methods explicitly disable momentum, as trajectory-averaged gradients can delay responsiveness during abrupt task transitions in sequential learning \cite{benjamin2024continual}. We also exclude the use of adaptive optimizers such as Adam \cite{kingma2014adam}. Since Adam applies dynamic element-wise preconditioning based on gradient moments, it continuously distorts the underlying geometry of the parameter space \cite{chng2025preconditioners}. This warping effectively severs the theoretical link between the optimizer's trajectory and the algorithm's curvature-based regularization, neutralizing the intended stability-plasticity trade-off \cite{dohare2024loss}. Models are trained for $50$ epochs on Split CIFAR-100, and $20$ epochs on both CORe50 and Split CIFAR-10. The sampling utilized in ER, RWalk, OCAR, and TMLN are implemented as a reservoir buffer \cite{vitter1985random} with $500$ samples.

\subsection{Main Results}
\label{subsec:results}
\begin{table}[ht]
    \centering
    \caption{Performance metrics on Split CIFAR-100 and CORe50. Final Accuracy (ACC), Average Forgetting (FM), and Intransigence (INT) are reported. \textit{Joint (oracle)} serves as an upper bound for accuracy. All methods are averaged over five runs.}
    \resizebox{\textwidth}{!}{%
    \begin{tabular}{lcccccc}
        \toprule
        & \multicolumn{3}{c}{Split CIFAR-100} & \multicolumn{3}{c}{CORe50} \\
        \cmidrule(lr){2-4} \cmidrule(lr){5-7}
        Method & ACC ($\uparrow$) & FM ($\downarrow$) & INT ($\downarrow$) & ACC ($\uparrow$) & FM ($\downarrow$) & INT ($\downarrow$) \\
        \midrule
        Joint (oracle) & $94.39 \pm 0.14$ & $--$ & $--$ & $80.03 \pm 0.00$ & $--$ & $--$ \\
        \midrule
        Naive SGD & $7.78 \pm 0.40$ & $66.53 \pm 2.40$ & $21.67 \pm 2.55$ & $63.72 \pm 0.57$ & $41.13 \pm 0.69$ & $23.88 \pm 0.00$ \\
        ER & $9.23 \pm 0.19$ & $73.42 \pm 3.37$ & $\bf{14.99 \pm 3.24}$ & $77.36 \pm 1.50$ & $23.89 \pm 1.68$ & $1.57 \pm 0.11$ \\
        EWC++ & $6.01 \pm 0.06$ & $58.18 \pm 0.59$ & $33.64 \pm 0.39$ & $76.94 \pm 0.52$ & $16.44 \pm 0.76$ & $9.51 \pm 0.32$ \\
        RWalk & $12.12 \pm 1.35$ & $70.01 \pm 1.46$ & $15.04 \pm 0.38$ & $67.76 \pm 0.68$ & $\bf{4.25 \pm 0.96}$ & $32.32 \pm 0.20$ \\
        UPGD & $8.50 \pm 1.32$ & $53.12 \pm 1.91$ & $34.77 \pm 3.38$ & $67.38 \pm 0.92$ & $33.57 \pm 0.93$ & $\bf{1.31 \pm 0.24}$ \\
        OCAR & $3.06 \pm 0.38$ & $\bf{30.80 \pm 1.98}$ & $62.60 \pm 2.31$ & $41.15 \pm 0.55$ & $10.18 \pm 0.65$ & $50.22 \pm 0.30$ \\
        \textbf{TMLN (Ours)}& $\bf{17.63 \pm 0.34}$ & $48.29 \pm 3.12$ & $29.11 \pm 2.74$ & $\bf{77.94 \pm 1.52}$ & $19.98 \pm 1.17$ & $3.64 \pm 0.72$ \\
        \bottomrule
    \end{tabular}
           }
    \label{tab:main_results}
\end{table}

Here, we present our results of our strategy for the Split CIFAR-100 and CORe50 experiments. As summarized in Table~\ref{tab:main_results}, TMLN consistently outperforms all non-oracle baselines in final average accuracy in both Split CIFAR-100 and CORe50. On the challenging Split CIFAR-100 benchmark, TMLN achieves a final accuracy of 17.63\%, a significant improvement over the strongest non-trajectory baseline, RWalk. 

The primary driver of this performance is the drastic reduction in average forgetting. TMLN limits forgetting to 48.29\%, compared to 58.18\% for EWC++ and 53.12\% for UPGD. This empirical evidence validates how TMLN preserves task-specific knowledge much more effectively than static regularization strategies. The higher intransigence of TMLN (of 29.11\%) compared to unconstrained baselines such as ER and naive SGD demonstrates the mathematical cost of utilizing preconditioning to prevent forgetting. However, by introducing dynamic resistance to rapid weight changes rather than rigidly constraining the loss landscape, TMLN maintains more plasticity than second-order methods such as OCAR and EWC++. OCAR’s failure to reach a meaningful accuracy suggests that its regularizing mechanism is too restrictive, over-constraining the parameters to a point where the model is structurally prevented from navigating toward the minima of new task distributions.

In the CORe50 dataset, the advantage of TMLN is even more pronounced. Our method achieves 77.94\% accuracy, representing a $\sim$14\% absolute improvement over Naive SGD. Crucially, the forgetting is reduced by more than half (from 41.13\% to 19.98\%), demonstrating that the trajectory-modulated geometric anchoring remains robust even as the number of tasks and domain shifts increases.

\begin{figure}[htpb!]
    \centering
    \includegraphics[width=1.1\textwidth]{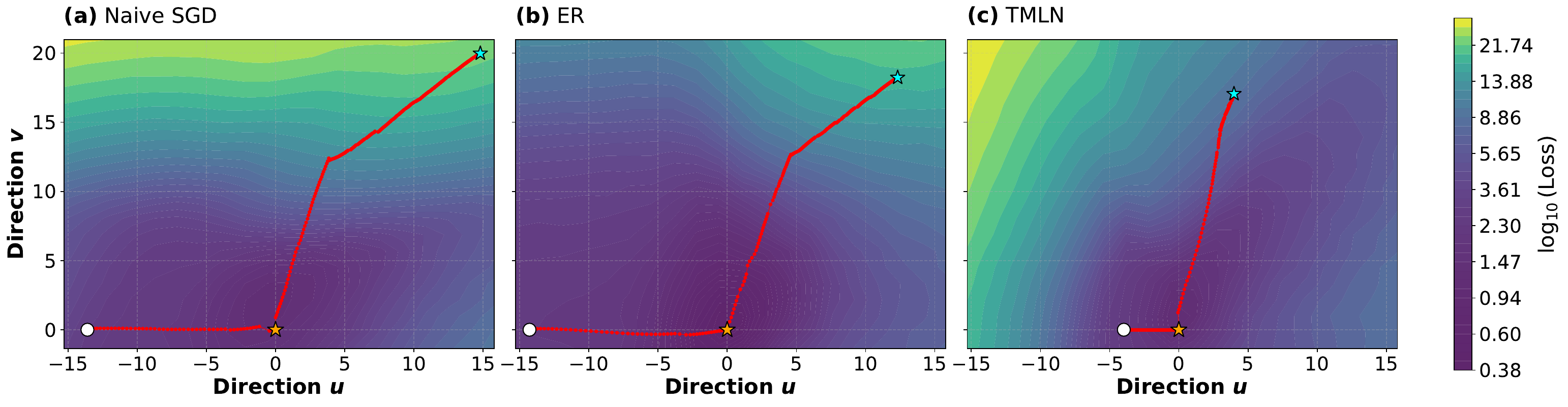}
    \caption{\textbf{Optimization trajectories during sequential task transition.} 
        In the Split CIFAR-100 task, network weight trajectories (shown in red) are projected onto a 2D subspace defined by the primary Task 1 optimization direction ($\boldsymbol{u}$) and the orthogonalized Task 2 update direction ($\boldsymbol{v}$). Background contours represent the $\log_{10}$ loss evaluated on Task 1. Markers denote initialization (white circle), the Task 1 minimum (orange star), and the Task 2 solution (cyan star). While \textbf{(a) Naive SGD} and \textbf{(b) Experience Replay (ER)} exhibit destructive parameter drift along the $\boldsymbol{u}$-axis, our proposed \textbf{(c) TMLN} strictly confines optimization to the orthogonal $\boldsymbol{v}$-subspace. Notably, TMLN conditions the network to converge to a wider, flatter Task 1 minimum, allowing substantial parameter displacement during new-task acquisition while safely remaining within the established low-energy functional basin.}
    \label{fig:fig2}
\end{figure}

\paragraph{Visualizing the Navigation Strategy.}
To provide geometric intuition for the success of our method, we visualize the sequential training process in a projected 2D parameter space. Following the visualization techniques in \cite{li2018visualizing}, we use filter-normalized random directions to plot the loss contou3we2rs of Task 1 of the Split CIFAR-100 setting. As illustrated in Fig.~\ref{fig:fig2}, the dotted lines trace the optimization path of the model parameters ($\theta$) as it learns Task 2, starting from model initialization.

The visualizations reveal the exact mechanism of catastrophic forgetting in baseline methods. For Naive SGD, the optimizer takes a near-linear Euclidean step from the safe Task 1 valley directly toward the Task 2 region. By blindly traversing the steepest loss contours of the previous task, this trajectory maximizes spatial interference, effectively climbing "out" of the Task 1 minimum. While ER slightly biases the update, its trajectory still aggressively crosses past-task boundaries to achieve plasticity.

In contrast, our method, TMLN, exhibits a fundamentally distinct optimization dynamic. The trajectory does not move directly toward Task 2. Instead, it is smoothly regularized by the Trajectory-Modulated FIM to curve along the low-loss valley floor of Task 1. This deterministic navigation allows the model to find a joint low-loss solution for both tasks without ever exiting the "safe zone" of past knowledge. This visual evidence supports our core claim: by smoothing updates along a geometrically anchored trajectory, TMLN intrinsically accounts for off-diagonal parameter interactions, preventing catastrophic interference while maximizing future plasticity.

\paragraph{Analyzing the Stability-Plasticity trade-off.}
A critical observation in Table~\ref{tab:main_results} is the behavior of the \textit{intransigence} metric. While TMLN exhibits higher intransigence than Naive SGD or RWalk, it achieves a significantly better balance than other second-order or utility-based methods like UPGD and EWC++. 

This suggests that while all stability-promoting methods introduce some degree of resistance to new learning, TMLN's navigation strategy is less obstructive than freezing regularization strategies. By allowing the network to move along the invariant sub-manifolds of past tasks, we maximize the remaining degrees of freedom available for new task acquisition, and provide a solution that Pareto-dominates traditional regularization strategies \cite{jedlicka2022contributions, pallasdies2021neural}. TMLN achieves the highest accuracy in the cohort despite the inherent cost of stability, effectively shifting the Pareto front of the stability-plasticity dilemma closer to the ideal joint-learning oracle. To strengthen these findings, a more comprehensive analysis of stability-plasticity tradeoffs is presented in Appendix~\ref{appendix:pareto} in the Split CIFAR-10 setting.

\subsection{Ablation and Analytical Studies}
\label{sec:ablation}
To evaluate the relative contributions of the Riemannian metric tensor and the trajectory modulation mechanism, we conduct an ablation study on Split CIFAR-10, comparing the full TMLN framework against two distinct variants and a standard SGD baseline (as shown in Table~\ref{tab:ablation}). 

To demonstrate the necessity of our dynamic path integral, we first ablate the trajectory component to create \textit{TMLN-Static}. In this variant, the preconditioning metric is determined entirely by the static FIM calculated at task boundaries, mimicking the discrete optimization geometry utilized in earlier curvature-based methods \cite{rattray1998natural, kirkpatrick2017overcoming}. While TMLN-Static significantly outperforms Standard SGD, it exhibits the highest intransigence ($14.56 \pm 4.14\%$) among all TMLN variants. This suggests that calculating geometry only at the task's end leads to a ``rigid'' regularization that over-constrains the model, making it difficult to learn subsequent tasks.

Additionally, to explicitly prove the necessity of navigating the true curvature of the loss landscape, we introduce \textit{TMLN-Euclidean}. This ablation strictly preserves our continuous trajectory modulation mechanism but completely strips away the Riemannian metric tensor ($\mathbf{F}$). Instead of preconditioning gradients with the landscape's intrinsic geometry, TMLN-Euclidean assumes a flat, isotropic parameter space. Mathematically, it replaces the curvature estimation with a purely Euclidean displacement penalty ($\|\Delta \theta\|^2$), reducing the trajectory integration to a normalized variant of the Synaptic Intelligence (SI) path integral \cite{zenke2017continual}. Interestingly, TMLN-Euclidean achieves a higher average accuracy than TMLN-Static, indicating that the temporal integration of the optimization path is a potent regularizer for stability. However, without the FIM to identify sensitive directions in the weight space, the model remains more stubborn when encountering new data.

The \textit{full TMLN} framework, which combines both FIM-based geometry and trajectory modulation, achieves the best overall performance of $51.55\%$. The most significant improvement is observed in its intransigence, which drops to $9.94\%$. This underscores the efficacy of our approach. By navigating the intrinsic geometry of the loss landscape rather than a flat Euclidean approximation, TMLN effectively minimizes the interference between tasks without sacrificing the capacity to learn new information.

\begin{table}[hbtp!]
\centering
\caption{Ablation results of TMLN. We evaluate the impact of the Riemannian metric tensor (FIM) and historical-trajectory modulation on Split CIFAR-10. Results are averaged over five runs.}
\label{tab:ablation}
\setlength{\tabcolsep}{6pt}
\renewcommand{\arraystretch}{1.15}
\newcolumntype{L}[1]{>{\RaggedRight\arraybackslash}p{#1}}
\newcolumntype{C}{>{\centering\arraybackslash}X}
\begin{tabularx}{\linewidth}{@{}L{2.6cm}CCCCC@{}}
\toprule
Method & Geometry (FIM) & Trajectory & ACC\newline($\uparrow$)& Forgetting\newline($\downarrow$) & Intransigence\newline($\downarrow$) \\
\midrule
Standard SGD & -- & -- & $19.04\pm0.15$ & $89.89\pm1.26$ & $0.00\pm0.48$ \\
TMLN-Static & $\checkmark$ & -- & $46.97\pm2.56$ & $41.39\pm4.30$ & $14.56\pm4.14$ \\
TMLN-Euclidean & -- & $\checkmark$ & $50.35\pm1.00$ & $\mathbf{39.48\pm1.85}$ & $12.27\pm0.78$ \\
\textbf{TMLN (Full)} & $\checkmark$ & $\checkmark$ & $\mathbf{51.55\pm0.56}$ & $40.32\pm3.67$ & $\mathbf{9.94\pm3.66}$ \\
\bottomrule
\end{tabularx}
\end{table}

\section{Discussion and Limitations}
\label{sec:disc}
While TMLN provides a robust geometric framework for continual learning, it requires certain mathematical compromises to achieve scalability. By relying on a diagonal approximation of the FIM, TMLN explicitly drops the off-diagonal curvature elements that capture true, instantaneous parameter covariances \cite{soen2024trade}. Methods that use the K-FAC model can interact with these more accurately but incur computational overhead. However, our empirical results defend this trade-off. By preconditioning the diagonal FIM with the historical optimization trajectory, TMLN naturally accounts for the curvature estimation over time and implicitly captures higher-order dependencies, providing a highly scalable $\mathcal{O}(N)$ alternative that competes with intractable second-order methods.

Additionally, while TMLN expertly mitigates catastrophic forgetting, it does not explicitly prevent the long-term "loss of plasticity" observed in deeply sequential learning \cite{dohare2024loss}. Traversing the Pareto frontier of stability and plasticity reveals a different mechanism from that of capacity: we are fundamentally facing a utility problem. It has recently been discovered that injecting noise can address this \cite{vastola2025gradient}. However, fully unlocking a network's long-term learning potential will require deeper structural shifts. Recent frameworks, such as Nested Learning \cite{behrouz2025nested}, argue towards multi-level optimization philosophies to avoid network vulnerability to "anterograde amnesia" \cite{scoville1957loss}, thus providing a critical direction for future work.

\section{Conclusion}
\label{sec:conclusion}
In this work, we argued that the core challenge of continual learning must be fundamentally recast from localized parameter penalization into a problem of optimal geometric navigation. We introduced TMLN to realize this and delivered both improved stability-plasticity balance and performance accuracy compared to baseline and state-of-the-art continual learning strategies. Through the integration of optimization-path-based importance scores into a unified geometric framework, we demonstrated the stability-plasticity dilemma as a geometric landscape to be optimally navigated rather than an insurmountable barrier.

\newpage
\bibliographystyle{unsrt} 
\bibliography{literature}


\newpage
\appendix

\section{Derivation of the TMLN Update Rule}
\label{appendix:derivation}

In Continual Learning (CL), the optimizer must navigate a highly non-convex loss landscape to minimize the loss of the current task while avoiding regions that drastically increase the loss of previous tasks. The Trajectory Modulated Landscape Navigator (TMLN) achieves this by dynamically reshaping the local gradient trust region under a task-conditioned diagonal metric. Unlike static penalty methods (e.g., EWC) that anchor weights to past optima, TMLN acts as a second-order preconditioner, utilizing both local buffer curvature and the historical parameter trajectory to guide optimization.

\subsection{The Proximal Optimization Framework}
At training step $t$, standard gradient descent can be viewed as solving a local linearized objective bounded by an isotropic Euclidean penalty to prevent taking overly large steps:
\begin{equation}
\label{eq:euclidean_penalty}
\Delta \theta_t = \arg\min_{\Delta \theta} \left( \nabla \mathcal{L}(\theta_t)^\top \Delta \theta + \frac{1}{2\eta} \|\Delta \theta\|_2^2 \right)
\end{equation}

The objective in Eq.~\ref{eq:euclidean_penalty} has the closed-form minimizer $\Delta\theta_t = -\eta \nabla_\theta \mathcal{L}(\theta_t)$, so the proximal view is exactly equivalent to standard gradient descent.

TMLN replaces the isotropic Euclidean distance with a specialized Mahalanobis distance that dynamically warps the parameter space. We define the TMLN surrogate objective as:
\begin{equation}\label{eq:mahalanobis_penalty}
J(\Delta \theta) = \underbrace{\nabla_\theta \mathcal{L}_{curr}(\theta_t)^\top \Delta \theta}_\text{First-order Alignment} + \underbrace{\frac{1}{2} \Delta \theta^\top \tilde{H}_t \Delta \theta}_\text{Mahalanobis Constraint}
\end{equation}
where $\tilde{H}_t$ is the \textbf{Trajectory-Modulated Metric Tensor}, a positive semi-definite matrix that scales the optimization steps. The first-order alignment guides the step toward the steepest descent of the current loss; the Mahalanobis-constrained penalty limits the step size based on the warped geometry defined by the trajectory.

\subsubsection{Mahalanobis Distance Derivation in the TMLN Surrogate Objective}
\label{sec:mahalanobis_derivation}

The squared Euclidan distance used in Eq.~\ref{eq:euclidean_penalty} is defined as such \cite{boyd2004convex,strang2016introduction,horn2017matrix}: 
\begin{equation}
\label{eq:Euclid_dist}
    d_E(\theta,\theta_t)^2:= ||\theta-\theta_t||^2_2 = (\theta-\theta_t)^\top I(\theta-\theta_t)
\end{equation}
where the identity matrix $I$ implies that the parameter space is isotropic \cite{amari1998natural}. 

The Mahalanobis distance is similar, but accounts for correlations and variances between variables \cite{demaesschalck2000mahalanobis,anderson2003introduction}. Given some positive semi-definite matrix $M$, representing the local curvature or geometry of the loss surface, the squared Mahalanobis distance is \cite{demaesschalck2000mahalanobis,anderson2003introduction,nocedal2006numerical}:
\begin{equation}
\label{eq:maha_dist}
    d_M(\theta,\theta_t)^2 := (\theta-\theta_t)^\top M(\theta-\theta_t)
\end{equation}
When $M \neq cI$ (for any scalar ($c>0$)), the induced geometry is anisotropic: distances (and therefore penalties) depend on direction, shrinking movement along high-curvature (large-eigenvalue) directions and allowing larger movement along low-curvature directions \cite{nocedal2006numerical,boyd2004convex}.

We now introduce the metric tensor $\tilde{H}_t\approx\eta^{-1}M_t$. This acts as a local distorter of the geometry. Unlike Euclidean distance that treats the parameter space as a flat sheet of paper, the Mahalanobis term treats it like a contoured landscape \cite{amari1998natural,absil2008optimization}. The entire metric tensor $\tilde{H}_t$ determines both the direction and the magnitude of the optimization step by acting as an inverse learning-rate-valued matrix, also known as a ``curvature matrix'' \cite{amari1998natural,lauritzen1996graphical}.

We can now map the original surrogate objective for the Euclidean space (defined in Eq.~\ref{eq:euclidean_penalty} and Eq.~\ref{eq:Euclid_dist}) to the transformed contoured space (defined in Eq.~\ref{eq:mahalanobis_penalty} and Eq.~\ref{eq:maha_dist}). By substituting $\Delta\theta = \theta-\theta_t$, using the matrix tensor $\tilde{H}_t$ identity, the objective penalty becomes:
\begin{equation}
    \dfrac{1}{2\eta} \|\Delta \theta\|_2^2 = \dfrac{1}{2\eta} d_E(\Delta\theta)^2 \;\;\;\;\longmapsto\;\;\;\; \dfrac{1}{2\eta}d_{\tilde{H}_t} (\Delta\theta)^2 = \dfrac{1}{2}\Delta\theta^\top \tilde{H}_t\Delta\theta \qed
\end{equation}
as required in Eq.~\ref{eq:mahalanobis_penalty}

\subsection{Constructing the Trajectory-Modulated Metric Tensor (\texorpdfstring{$\tilde{H}_t$}{Ht})}\label{sec:TMMT_construction}
To prevent catastrophic forgetting while allowing plasticity, TMLN constructs $\tilde{H}_t$ from three distinct components:

\begin{enumerate}
    \item \textbf{Local Memory Curvature ($F_{emp}$):} While the True Fisher Information Matrix defines the exact Riemannian metric tensor, it is often intractable to compute continuously \cite{van2025computation}. Instead, we use the Empirical Fisher computed over a replay buffer $\mathcal{M}$. The Empirical Fisher serves as a practical, computationally efficient upper bound on the gradient covariance, penalizing movement in parameters that exhibit high gradient variance on the buffered examples:
    \begin{equation}
    F_{emp}(\theta) = \frac{1}{|\mathcal{M}|} \sum_{(x,y) \in \mathcal{B}} \nabla_\theta \ell(f(x; \theta), y) \nabla_\theta \ell(f(x; \theta), y)^\top
    \end{equation}
    
    \item \textbf{Path Integral Modulation ($S$):} To capture long-term historical importance, we adapt the trajectory-tracking concepts introduced in Synaptic Intelligence (SI) \cite{zenke2017continual} and Riemannian Walk (RWalk) \cite{chaudhry2018riemannian}. We accumulate the path integral of the gradients along the optimization trajectory:
    \begin{equation}
    \omega_k = \sum_{t \in \text{task } k} -\nabla \mathcal{L}(\theta_t) \cdot \Delta \theta_t
    \end{equation}
    Following the normalization strategy of RWalk, we divide this path integral by the change in the buffer's parameter sensitivity to yield the importance matrix $S$, defined element-wise for parameter $i$ as:
    \begin{equation}
    S_i = \frac{\max(0, \omega_{k, i})}{\frac{1}{2} F_{emp, ii} (\Delta \theta_i)^2 + \epsilon}
    \end{equation}
    
    \item \textbf{Isotropic Damping ($\lambda I$):} A base Tikhonov regularization term ensures $\tilde{H}_t$ remains strictly positive definite (invertible) and guarantees a baseline learning rate globally. This is important to prevent ill-conditioned (or singular) curvature estimates from producing unstable, excessively large updates \cite{martens2020new}, making the preconditioned step $\Delta\theta_t^\star=-\tilde{H}_t^{-1}\nabla_\theta \mathcal{L}_{\mathrm{curr}}(\theta_t)$ well-defined at every iteration (derivation of the preconditioned step can be found in Appendix \ref{sec:quadratic_surrogate_minimizer}).
\end{enumerate}

Combining these, the full TMLN metric tensor is defined as:
\begin{equation}
\tilde{H}_t = F_{emp}(\theta_t) + \lambda I + \gamma \text{diag}(S)
\end{equation}
Here, the hyperparameter $\gamma$ controls the strength of the historical trajectory modulation, and both $\lambda$ and $\gamma$ absorb the necessary physical units to ensure dimensional consistency when summing the covariance, identity, and importance score matrices.

\subsubsection{Closed-form minimizer of the Quadratic Surrogate Objective}
\label{sec:quadratic_surrogate_minimizer}

Before linking the composite construction of $\tilde{H}_t$ to the $\eta^{-1}M_t$ decomposition, we first record the closed-form minimizer of the quadratic surrogate objective used throughout this appendix. Starting from Eq.~\eqref{eq:mahalanobis_penalty},
\begin{equation}
	J(\Delta\theta)
	=
	\nabla_\theta \mathcal{L}_{\mathrm{curr}}(\theta_t)^\top \Delta\theta
	+
	\frac{1}{2}\Delta\theta^\top \tilde{H}_t \Delta\theta,
\end{equation}
we compute the gradient with respect to $\Delta\theta$:
\begin{equation}
	\nabla_{\Delta\theta} J(\Delta\theta)
	=
	\nabla_\theta \mathcal{L}_{\mathrm{curr}}(\theta_t)
	+
	\tilde{H}_t \Delta\theta,
\end{equation}
where $\tilde{H}_t$ is taken to be symmetric as a quadratic penalty matrix. The first-order optimality condition for an unconstrained differentiable objective is
\begin{equation}
	\nabla_{\Delta\theta} J(\Delta\theta^\star) = 0
	\quad\Longleftrightarrow\quad
	\nabla_\theta \mathcal{L}_{\mathrm{curr}}(\theta_t) + \tilde{H}_t \Delta\theta^\star = 0.
\end{equation}
Assuming $\tilde{H}_t$ is invertible (e.g., $\tilde{H}_t \succ 0$ due to isotropic damping with $\lambda>0$), we obtain the closed-form minimizer
\begin{equation}
	\Delta\theta_t^\star
	=
	-\tilde{H}_t^{-1}\nabla_\theta \mathcal{L}_{\mathrm{curr}}(\theta_t).
	\label{eq:tmln_precond_update}
\end{equation}

\subsubsection{Linking the Composite Metric \texorpdfstring{$\tilde{H}_t$} \\ to the \texorpdfstring{$\eta^{-1}M_t$}\\
Decomposition}
\label{sec:Ht_eta_inverse_M_link}

In Appendix~\ref{sec:mahalanobis_derivation}, we introduced the decomposition
\begin{equation}
	\tilde{H}_t \equiv \eta^{-1} M_t,
	\label{eq:Ht_eta_inv_M_def}
\end{equation}
which separates the \emph{scalar} learning-rate factor $\eta$ from a \emph{dimensionless} anisotropic modulator $M_t$.

In the full TMLN construction, the trajectory-modulated metric is defined as
\begin{equation}
	\tilde{H}_t
	= F_{\mathrm{emp}}(\theta_t) + \lambda I + \gamma\,\mathrm{diag}(S).
	\label{eq:Ht_composite}
\end{equation}
To make Eq.~\eqref{eq:Ht_composite} consistent with Eq.~\eqref{eq:Ht_eta_inv_M_def}, we calibrate the isotropic damping such that the method reduces to vanilla gradient descent in the isotropic limit. Specifically, if $F_{\mathrm{emp}}(\theta_t)=\mathbf{0}$ and $\gamma=0$, then $\tilde{H}_t=\lambda I$ and the update $\Delta\theta_t^\star=-\tilde{H}_t^{-1}\nabla_\theta \mathcal{L}_{\mathrm{curr}}(\theta_t)$ (from Eq.~\ref{eq:tmln_precond_update}) reduces to $\Delta\theta_t^\star = -(1/\lambda)\nabla_\theta \mathcal{L}_{\mathrm{curr}}(\theta_t)$ \cite{boyd2004convex,nocedal2006numerical}. Setting
\begin{equation}
	\lambda := \eta^{-1}
	\label{eq:lambda_eta_inv}
\end{equation}
therefore recovers 
\begin{equation}
\Delta\theta_t^\star = -\eta\nabla_\theta \mathcal{L}_{\mathrm{curr}}(\theta_t)
\end{equation}
in the isotropic limit.

Substituting Eq.~\eqref{eq:lambda_eta_inv} into Eq.~\eqref{eq:Ht_composite} and factoring out $\eta^{-1}$ yields
\begin{align}
	\tilde{H}_t
	&= \eta^{-1} I + F_{\mathrm{emp}}(\theta_t) + \gamma\,\mathrm{diag}(S) \nonumber\\
	&= \eta^{-1}\underbrace{\left(I + \eta F_{\mathrm{emp}}(\theta_t) + \eta\gamma\,\mathrm{diag}(S)\right)}_{=:~M_t}.
	\label{eq:Ht_factored}
\end{align}
Comparing Eq.~\eqref{eq:Ht_factored} with Eq.~\eqref{eq:Ht_eta_inv_M_def} identifies the anisotropic modulator as
\begin{equation}
	M_t := I + \eta F_{\mathrm{emp}}(\theta_t) + \eta\gamma\,\mathrm{diag}(S).
	\label{eq:Mt_definition}
\end{equation}
Thus, $\eta^{-1}$ provides the baseline isotropic quadratic penalty, while $M_t$ captures state- and task-dependent departures from Euclidean geometry \cite{amari1998natural,boyd2004convex, horn2017matrix}. Recalling from Appendix~\ref{sec:mahalanobis_derivation}, when $M_t \neq cI$ (for any scalar $c>0$), the induced quadratic penalty is anisotropic, and the effective stepsize becomes direction-dependent through $\tilde{H}_t^{-1}$.

\subsection{Solving for the Preconditioned Update Rule}
Starting from the closed-form minimizer in Eq.~\eqref{eq:tmln_precond_update},
\begin{equation}
\Delta\theta_t^\star
= -\tilde{H}_t^{-1}\nabla_\theta \mathcal{L}_{\mathrm{curr}}(\theta_t),
\end{equation}
and substituting the composite definition $\tilde{H}_t = F_{\mathrm{emp}}(\theta_t) + \lambda I + \gamma\,\mathrm{diag}(S)$, we obtain the explicit update:
\begin{equation}
\Delta\theta_t^\star
= -\left(F_{\mathrm{emp}}(\theta_t) + \lambda I + \gamma\,\mathrm{diag}(S)\right)^{-1}
\nabla_\theta \mathcal{L}_{\mathrm{curr}}(\theta_t).
\end{equation}

\subsection{The Diagonal Tractability Relaxation}
In modern deep neural networks, instantiating and inverting the full matrix $F_{emp}$ requires $\mathcal{O}(N^2)$ space and $\mathcal{O}(N^3)$ compute. While block-diagonal approximations (e.g., K-FAC) can capture layer-wise parameter correlations, the strict memory limitations and latency requirements of our specific online continual learning setting necessitate an even lighter footprint. 

Therefore, TMLN applies a strict diagonal relaxation. We restrict $F_{emp}$ to its diagonal, $\text{diag}(F_{emp})$. Because the sum of diagonal matrices is diagonal, the inversion becomes a trivial element-wise division. For a single parameter $\theta_i$, the update rule simplifies to the highly scalable formula utilized in our implementation:
\begin{equation}
(\Delta \theta)_i = - \frac{1}{F_{emp, ii} + \lambda + \gamma S_i} \left( \nabla \mathcal{L}_{curr}(\theta_t) \right)_i
\end{equation}

\section{Experimental Details and Hyperparameters}
\label{app:hyperparameters}

\subsection{Dataset Augmentation}
Across all experiments (Split CIFAR-100/10 and CORe50), we applied standard data augmentation to the training sets prior to optimization.

\begin{itemize}
    \item \textbf{CIFAR:} Each image is zero-padded by 4 pixels on all sides and randomly cropped back to its original $32 \times 32$ spatial resolution. Next, a random horizontal flip is applied with a probability of $0.5$. Finally, we apply a color jittering transformation that randomly perturbs the image brightness by a factor chosen uniformly from $[-63/255, +63/255]$. 
    \item \textbf{CORe50:} Each image is subjected to a random resized crop, which extracts a random patch of the original image and resizes it to $224 \times 224$ pixels (utilizing the default PyTorch scale and aspect ratio bounds). This is followed by a random horizontal flip ($p=0.5$) and a brightness perturbation identical to the CIFAR-10 setup, where the brightness factor is jittered by a factor chosen uniformly from $[-63/255, +63/255]$.
\end{itemize}

\subsection{Hyperparameter Specifications}
The following table outlines the specific coefficients used for each method in Split CIFAR-100. These were selected via grid search. 

\begin{table}[h]
\centering
\caption{Hyperparameter specifications for baselines and the proposed method.}
\label{tab:hyperparameters}
\begin{tabular}{@{}lll@{}}
\toprule
Method & Parameter & Value \\ \midrule
\multirow{2}{*}{\textbf{TMLN (Ours)}} & $\lambda$ & $0.1$ \\ 
 & $\gamma$ & $1.0$ \\
\midrule
ER & $\mathcal{B}$ & $500$ \\ \midrule
EWC++ & $\lambda$ & $10000$ \\ \midrule
\multirow{3}{*}{RWalk} & $\lambda$ & $10$ \\
 & $\alpha$ & $0.8$ \\
 & $\Delta t$ & $50$ \\ \midrule
\multirow{2}{*}{UPGD} & $\beta_{utility}$ & $0.99$ \\
 & $\sigma$ & $0.1$ \\ \midrule
OCAR & $\tau$ & $0.1$ \\
\bottomrule
\end{tabular}
\end{table}

\section{Computational and Memory Overhead}
\label{appendix:overhead}

A critical consideration for any continuous learning system, particularly those leveraging second-order optimization or Riemannian geometry, is the computational and memory footprint required to maintain stability. While ER sustains a memory cost tied to the raw data distribution, regularization- and optimization-based methods (EWC, OCAR, TMLN) have overhead tied to the network architecture and the geometric state of past tasks. 

Table \ref{tab:overhead} details the empirical resource consumption of all methods evaluated on the CIFAR-100 benchmark. All timing and memory evaluations were conducted on a single Tesla P100 GPU using a consistent batch size of $128$.

\begin{table}[htp!]
\centering
\caption{Computational and Memory Overhead on CIFAR-100. Time is normalised relative to Naive SGD. Auxiliary memory indicates the additional state required beyond the standard model parameters ($\theta$).}
\label{tab:overhead}
\begin{threeparttable}
\setlength{\tabcolsep}{6pt}
\renewcommand{\arraystretch}{1.15}

\newcolumntype{Y}{>{\RaggedRight\arraybackslash}X}

\begin{tabularx}{\linewidth}{@{}lYcc@{}}
\toprule
Method & Auxiliary Memory Form & Peak VRAM (GB) & Relative Training Time \\
\midrule
Naive & None ($0$) & 1.04 & 1.00x \\
ER (M=500) & Raw images ($\sim 6$ MB) & 1.94 & 1.04x \\
EWC++ & Diagonal Fisher ($\lVert\theta\rVert$) & 1.16 & 1.34x \\
OCAR &
K-FAC matrices ($\sum_{l=1}^{L}\big((d_{\mathrm{in}}^{(l)})^{2} + (d_{\mathrm{out}}^{(l)})^{2}\big)^{\ast}$) & 3.55 & 16.86x \\
\textbf{TMLN (Ours)} &
\textbf{Diagonal Fisher; stored $\theta_{\mathrm{prev}}$ ($2\times\lVert\theta\rVert$)} &
2.07 & 1.07x \\
\bottomrule
\end{tabularx}

\begin{tablenotes}[flushleft]
\footnotesize
\item[$\ast$] For OCAR, $L$ is the total number of layers in the network, $d_{\mathrm{in}}^{(l)}$ is the input dimension of layer $l$, and $d_{\mathrm{out}}^{(l)}$ is the output dimension of layer $l$.
\end{tablenotes}
\end{threeparttable}
\end{table}

\paragraph{Complexity Analysis of TMLN.} 
While TMLN incurs a computational penalty of 1.07x relative to Naive SGD, this overhead is fundamentally highly scalable. The structural navigation of TMLN relies on computing a diagonal approximation of the Fisher Information Matrix accompanied by a path integral over the optimization trajectory. Unlike exact second-order methods or sub-manifold projections that require computationally expensive matrix inversions or eigen-decompositions ($\mathcal{O}(d^3)$), both the diagonal Fisher and the path integral rely strictly on element-wise operations. Consequently, the time complexity per optimization step remains linearly bounded at $\mathcal{O}(|\theta|)$, where $|\theta|$ is the total number of network parameters. Furthermore, the auxiliary memory footprint is highly efficient; it requires storing strictly constant-size importance matrices (the Fisher diagonal and path integral tracking states), scaling at $\mathcal{O}(|\theta|)$ independent of the number of sequential tasks. This ensures resource consumption remains bounded well within modern hardware constraints without requiring privacy-violating raw data storage.

\section{Supplementary Loss Values for Trajectory Visualization}
\label{appendix:traj_loss}

\begin{table}[htpb!]
\centering
\caption{Optimization trajectory loss values across sequential tasks for Tasks 1 \& 2. Init, End of Task 1, and End of Task 2 columns correspond to the loss landscape visualization markers. Best final loss is in bold.}
\label{tab:loss_trajectory}
\begin{tabular}{lcccccc}
\toprule
 & \multicolumn{3}{c}{Task 1 Loss} & \multicolumn{3}{c}{Task 2 Loss} \\
\cmidrule(lr){2-4} \cmidrule(lr){5-7}
Method & Init & End of Task 1 & End of Task 2 & Init & End of Task 1 & End of Task 2 \\
\midrule
Naive SGD & 2.9338 & 1.3871 & 24.5702 & 11.1482 & 12.1247 & 6.3916 \\
ER        & 2.3030 & 0.3775 & 22.5264 & 10.4751 & 13.3639 & 5.2835 \\
\textbf{TMLN}      & 2.4773 & 1.1380 & \textbf{9.0663} & 8.7170 & 10.6379 & \textbf{1.0983} \\
\bottomrule
\end{tabular}
\end{table}

Here, we present supplementary data for Fig.~\ref{fig:fig2}. Loss values at checkpoints during the sequential training process are presented. In Table ~\ref{tab:loss_trajectory}, TMLN consistently has the lowest loss values for each task at the end of Task 2 ($9.0663$ and $1.0983$), demonstrating preservation of the knowledge of Task 1 during the learning of Task 2.

\section{Stability-Plasticity Trade-offs and Pareto Optimality}
\label{appendix:pareto}
Here, we provide a rigorous formalization of the stability-plasticity trade-off space, detail the hyperparameter sweeps required to generate the Pareto front, and offer a geometric interpretation of TMLN's Pareto optimality.

\subsection{Mathematical Formulation of the Trade-off Space}
\label{appendix:pareto_math}

To construct a meaningful Pareto frontier, we decouple average accuracy into its two opposing operational metrics: \textbf{Forgetting ($F$)} representing stability failure, and \textbf{Intransigence ($I$)} representing plasticity failure.

Following Chaudhry et al. \cite{chaudhry2018riemannian}, let $a_{k,j}$ be the accuracy of the model on task $j$ after learning task $k$. We define the trade-off axes as follows:

\begin{itemize}
    \item \textbf{Forgetting ($F_k$):} The average degradation in performance on past tasks.
    \begin{equation}
        F_k = \frac{1}{k-1} \sum_{j=1}^{k-1} \left( \max_{l \in \{1, \dots, k-1\}} a_{l,j} - a_{k,j} \right)
    \end{equation}
    
    \item \textbf{Intransigence ($I_k$):} The inability to acquire new knowledge compared to a joint-training oracle. Let $a_k^*$ be the accuracy of a model trained jointly on all data up to task $k$.
    \begin{equation}
        I_k = a_k^* - a_{k,k}
    \end{equation}
\end{itemize}

A solution $\theta_A$ is said to be Pareto optimal over a solution $\theta_B$ if $F(\theta_A) \leq F(\theta_B)$ and $I(\theta_A) \leq I(\theta_B)$, with at least one inequality being strict \cite{jedlicka2022contributions}. The \textbf{Pareto frontier} consists of the set of non-dominated configurations for a given algorithmic class. The optimal region sits in the bottom left area, where both stability and plasticity are sustained.

\subsection{Frontier Generation Methodology}
\label{appendix:pareto_methodology}

To construct the empirical Pareto frontiers presented in Fig.~\ref{fig:fig3}, we performed exhaustive grid searches over the primary stability-plasticity control variables for each method:

\begin{itemize}
    \item \textbf{Naive:} learning rate $\eta \in \{0.001, 0.005, 0.01, 0.05, 0.1\}$.
    \item \textbf{ER:} buffer size $\mathcal{B} \in \{100, 200, 500, 1000, 2000\}$.
    \item \textbf{EWC++ \& RWalk:} regularization coefficient
    \\ $\lambda \in \{100, 500, 1000, 5000, 10000\}$.
    \item \textbf{UPGD:} utility momentum $\beta_u \in \{0.9, 0.95, 0.99, 0.999, 0.9999\}$.
    \item \textbf{OCAR:} Tikhonov regularization parameter $\tau \in \{10^{-4}, 10^{-3}, 10^{-2}, 10^{-1}, 10\}$.
\end{itemize}

and compared them to our method, \textbf{TMLN}.

By plotting $(I_k, F_k)$ for all configurations and extracting the non-dominated points, we establish the true operational limits of each baseline.

\subsection{Analysis of Empirical Pareto Dominance}
\label{appendix:pareto_analysis}
\begin{table}[hbtp!]
\centering
\caption{Continual learning metrics for Split CIFAR-10 at various hyperparameter configurations.}
\label{tab:pareto}
\begin{tabular}{@{}ll rrr@{}}
\toprule
Method & Configuration & ACC ($\uparrow$) & FM ($\downarrow$) & INT ($\downarrow$) \\ 
\midrule

\multirow{1}{*}{Joint (oracle)} 
& $--$    & $75.98\pm0.56$ & $--$ & $--$ \\
\midrule
\multirow{5}{*}{Naive} 
& $\eta = 0.001$ & $19.04 \pm 0.25$ & $92.15 \pm 1.94$ & $-1.55 \pm 1.64$ \\
& $\eta = 0.005$ & $19.18 \pm 0.05$ & $91.28 \pm 0.80$ & $-0.74 \pm 0.89$ \\
& $\eta = 0.01$  & $19.04 \pm 0.15$ & $89.89 \pm 1.26$ & $0.00 \pm 0.48$ \\
& $\eta = 0.05$  & $16.78 \pm 3.39$ & $86.77 \pm 1.70$ & $5.09 \pm 4.93$ \\
& $\eta = 0.1$   & $12.14 \pm 3.70$ & $84.00 \pm 0.69$ & $13.92 \pm 4.13$ \\ 
\midrule

\multirow{5}{*}{ER} 
& $\mathcal{M} = 100$  & $16.32 \pm 0.82$ & $67.60 \pm 8.34$ & $24.49 \pm 9.32$ \\
& $\mathcal{M} = 200$  & $19.39 \pm 0.18$ & $82.03 \pm 1.20$ & $7.56 \pm 1.72$ \\
& $\mathcal{M} = 500$  & $18.90 \pm 2.02$ & $85.73 \pm 2.02$ & $5.08 \pm 3.33$ \\
& $\mathcal{M} = 1000$ & $18.20 \pm 1.95$ & $80.48 \pm 2.90$ & $10.12 \pm 2.94$ \\
& $\mathcal{M} = 2000$ & $28.35 \pm 4.68$ & $74.15 \pm 4.84$ & $4.44 \pm 1.99$ \\ 
\midrule

\multirow{5}{*}{EWC++} 
& $\lambda = 100$   & $18.63 \pm 0.30$ & $89.76 \pm 0.52$ & $1.83 \pm 0.47$ \\
& $\lambda = 500$   & $18.41 \pm 0.11$ & $88.12 \pm 0.76$ & $3.78 \pm 0.89$ \\
& $\lambda = 1000$  & $17.45 \pm 0.20$ & $83.31 \pm 1.27$ & $9.54 \pm 1.03$ \\
& $\lambda = 5000$  & $16.39 \pm 0.10$ & $76.15 \pm 0.31$ & $17.88 \pm 0.41$ \\
& $\lambda = 10000$ & $15.66 \pm 0.12$ & $73.38 \pm 0.94$ & $21.59 \pm 0.81$ \\ 
\midrule

\multirow{5}{*}{RWalk} 
& $\lambda = 100$   & $39.46 \pm 3.25$ & $61.56 \pm 5.08$ & $3.88 \pm 1.37$ \\
& $\lambda = 500$   & $38.75 \pm 2.50$ & $61.09 \pm 3.55$ & $5.31 \pm 1.53$ \\
& $\lambda = 1000$  & $34.94 \pm 1.51$ & $65.65 \pm 2.92$ & $5.47 \pm 2.08$ \\
& $\lambda = 5000$  & $32.42 \pm 1.61$ & $61.22 \pm 2.15$ & $13.00 \pm 0.66$ \\
& $\lambda = 10000$ & $30.28 \pm 2.07$ & $58.41 \pm 2.70$ & $18.26 \pm 0.76$ \\ 
\midrule

\multirow{5}{*}{UPGD} 
& $\beta_{u} = 0.9$    & $15.21 \pm 1.97$ & $81.72 \pm 1.77$ & $10.13 \pm 4.05$ \\
& $\beta_{u} = 0.95$   & $18.04 \pm 1.08$ & $85.83 \pm 3.67$ & $1.48 \pm 0.85$ \\
& $\beta_{u} = 0.99$   & $18.21 \pm 0.55$ & $87.94 \pm 2.29$ & $0.47 \pm 1.50$ \\
& $\beta_{u} = 0.999$  & $18.68 \pm 0.37$ & $88.28 \pm 1.01$ & $0.63 \pm 1.30$ \\
& $\beta_{u} = 0.9999$ & $17.99 \pm 1.27$ & $88.27 \pm 1.92$ & $0.95 \pm 1.67$ \\ 
\midrule

\multirow{5}{*}{OCAR} 
& $\tau = 10^{-4}$      & $14.34 \pm 0.00$ & $73.35 \pm 0.00$ & $20.73 \pm 0.00$ \\
& $\tau = 10^{-3}$      & $13.45\pm 0.00$ & $68.49\pm 0.00$ & $26.50\pm 0.00$ \\
& $\tau = 10^{-2}$      & $13.69\pm 0.02$ & $62.28\pm 0.05$ & $26.28\pm 0.04$ \\
& $\tau = 1$            & $14.71 \pm 0.00$ & $72.95 \pm 0.00$ & $20.75 \pm 0.00$ \\
& $\tau = 10$           & $13.85 \pm 0.00$ & $62.26 \pm 0.00$ & $30.20 \pm 0.00$ \\ 
\midrule

\multirow{5}{*}{\textbf{TMLN}} 
& $\gamma = 10^{-2}$      & $47.11 \pm 0.93$ & $\bf{36.34 \pm 10.67}$ & $19.49 \pm 11.18$ \\
& $\gamma = 10^{-1}$      & $46.49\pm1.08$ & $46.32 \pm4.29$ & $9.70\pm 3.05$ \\
& $\gamma = 1$    & $\bf{51.55 \pm 1.00}$ & $40.32 \pm 3.67$ & $9.94 \pm 3.66$ \\
& $\gamma = 10^{1}$            & $48.44\pm1.07$ & $43.88\pm2.32$ & $9.13\pm2.36$ \\
& $\gamma = 10^{2}$           & $40.91\pm2.69$ & $53.18\pm4.09$ & $7.33\pm1.96$ \\ 


\bottomrule
\end{tabular}
\end{table}

\begin{figure}[htp!]
    \centering
    \includegraphics[width=0.8\textwidth]{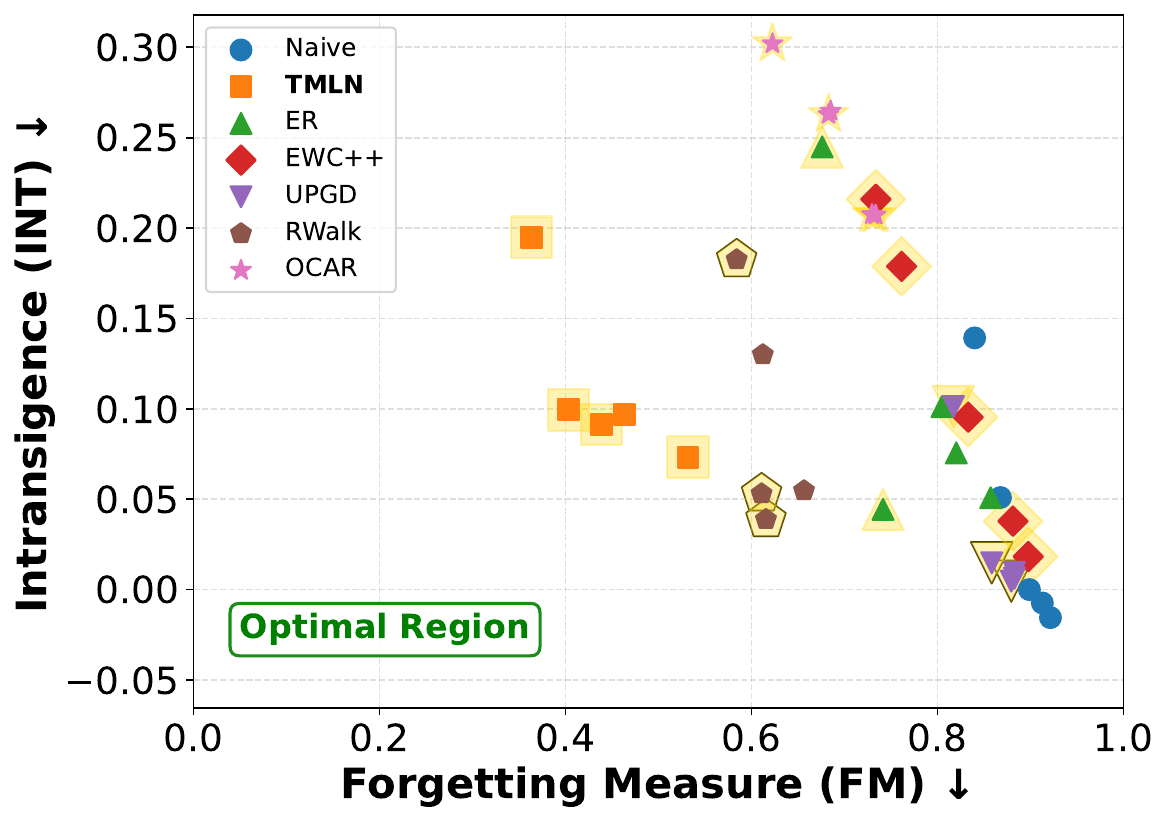}
    \caption{\textbf{Pareto Frontier of Stability-Plasticity Tradeoff.} Evaluation of Forgetting Measure (FM) vs. Intransigence (INT) across varied hyperparameter sweeps of CL strategies. Each marker represents individual configuration runs. Large yellow outlines delineate the local, non-dominated Pareto front for respective baseline methods. Black outlines denote the global optimal skyline. Our proposed method, TMLN (orange square), demonstrates state-of-the-art memory retention without severe loss of plasticity, unlike traditional baselines.}
    \label{fig:fig3}
\end{figure}

As shown in the resulting frontier plot in Fig.~\ref{fig:fig3}, traditional regularization and replay baselines exhibit severely sub-optimal trade-offs, struggling to simultaneously minimize forgetting and intransigence. In stark contrast, our proposed TMLN method strictly dominates the entire pareto space, independently forming below the optimal global skyline in the bottom-left optimal region. By achieving state-of-the-art memory retention without sacrificing plasticity, TMLN empirically proves its capacity to resolve the stability-plasticity dilemma where previous approaches struggle.

TMLN fundamentally alters this curve. Rather than navigating along the sub-optimal frontiers of the baselines, TMLN shifts the entire frontier inward toward the origin. Presented in Table ~\ref{tab:pareto}, we achieve an intransigence of $9.94\%$ while maintaining a Forgetting measure of $40.32\%$. For EWC++ to achieve a comparable intransigence score ($\sim 9.54\%$ at $\lambda=1000$), it suffers a catastrophic Forgetting rate of $83.31\%$. Therefore, TMLN optimally balances the trade-off space.

\subsection{Geometric Interpretation of the Frontier Shift}
\label{appendix:pareto_geometry}

The mechanism by which TMLN bypasses the traditional Pareto limits can be further understood through the geometry of the loss landscape.

Standard regularization (e.g., EWC) achieves stability by imposing an isotropic or diagonally-approximated quadratic penalty around the previous parameter state $\theta_{k-1}$. This effectively shrinks the "trust region'" or the available degrees of freedom for learning task $k$, strictly tying stability to structural rigidity (high intransigence).

TMLN, by contrast, restricts updates to the invariant sub-manifolds of past tasks. By utilizing historical and spatially aware preconditioning, TMLN ensures that parameter updates $\Delta \theta$ lie in the null space of the curvature matrix of previous tasks. Mathematically, this decouples the stability objective from the plasticity objective. Stability is maintained strictly through geometric orthogonality rather than magnitude penalization, leaving the maximal unconstrained subspace available for navigating the loss landscape of the new task. This structural advantage is what manifests empirically as a Pareto frontier situated significantly closer to the joint-learning oracle.


\end{document}